\documentclass[11pt]{article}

\usepackage{EMNLP2022}

\usepackage{times}
\usepackage{latexsym}

\usepackage[T1]{fontenc}
\usepackage[utf8]{inputenc}

\usepackage{microtype}

\usepackage{inconsolata}
\title{HIT-SCIR at MMNLU-22: Consistency Regularization for Multilingual Spoken Language Understanding}
\usepackage{graphicx}
\usepackage{subfigure}
\usepackage{CJKutf8}
\usepackage{multirow}
\usepackage{booktabs}
\usepackage{amsmath}
\usepackage{CJKutf8}

\author{Bo Zheng,~~Zhouyang Li,~~Fuxuan Wei,~~Qiguang Chen,~~Libo Qin,~~Wanxiang Che\thanks{\ \ Email corresponding.}\\
Harbin Institute of Technology \\
\texttt{\{bzheng,zhouyangli,fxwei,qgchen,lbqin,car\}@ir.hit.edu.cn}  \\}

\begin{document}
\maketitle
\begin{abstract}
Multilingual spoken language understanding (SLU) consists of two sub-tasks, namely intent detection and slot filling.
To improve the performance of these two sub-tasks, we propose to use consistency regularization based on a hybrid data augmentation strategy.
The consistency regularization enforces the predicted distributions for an example and its semantically equivalent augmentation to be consistent. 
We conduct experiments on the MASSIVE dataset under both full-dataset and zero-shot settings.
Experimental results demonstrate that our proposed method improves the performance on both intent detection and slot filling tasks. 
Our system\footnote{The code will be available at \url{https://github.com/bozheng-hit/MMNLU-22-HIT-SCIR}.} ranked 1st in the MMNLU-22 competition under the full-dataset setting.

\end{abstract}

\section{Introduction}
\begin{figure}[t]
    \centering
    \includegraphics[scale=0.43]{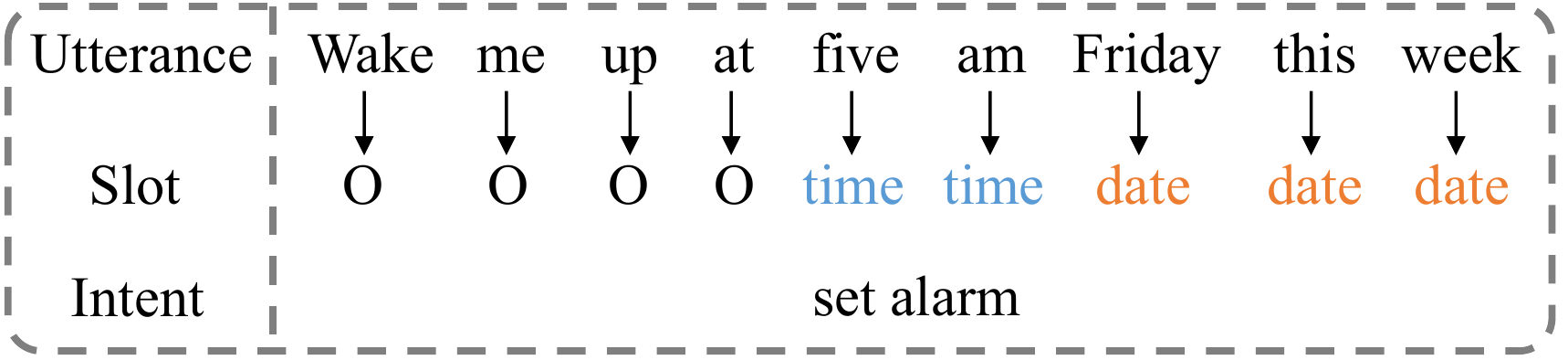}
    \caption{An English example from the MASSIVE dataset. The slot label `O' stands for the `Other' label.}
    \label{fig:example}
\end{figure}

The MMNLU-22 evaluation focuses on the problem of multilingual natural language understanding. 
It is based on the MASSIVE dataset~\cite{fitzgerald2022massive}, a multilingual spoken language understanding (SLU) dataset with two sub-tasks, including \textit{intent detection} and \textit{slot filling}. Specifically, given a virtual assistant utterance in an arbitrary language, the model is designed to predict the corresponding intent label and extract the slot results. An English example is illustrated in Figure~\ref{fig:example}.

Fine-tuning pre-trained cross-lingual language models allows task-specific supervision to be shared and transferred across languages~\citep{xlm,xlmr,mt5}. 
This motivates the two setting for the MMNLU-22 evaluation, namely the \textit{full-dataset} setting and the \textit{zero-shot} setting. 
Participants are allowed to use training data in all languages under the full-dataset setting, while they can only access the English training data under the zero-shot setting. 
The latter is also called zero-shot cross-lingual SLU in previous work~\citep{cosda-ml, qin-etal-2022-gl}.

Cross-lingual data augmentation methods have been proven effective to improve cross-lingual transferability, e.g., code-switch substitution~\citep{cosda-ml} and machine translation~\citep{xlm, xlda}. Most previous work directly utilizes the data augmentations as additional training data for fine-tuning. However, they ignore the inherent correlation between the original example and its semantically equivalent augmentation, which can be fully exploited with the \textit{consistency regularization}~\citep{xtune}. The consistency regularization enforces the model predictions to be more consistent for semantic-preserving augmentations.

Motivated by this, we propose to apply consistency regularization based on a hybrid data augmentation strategy, including data augmentation of machine translation and subword sampling~\citep{DBLP:conf/acl/Kudo18}. We use machine translation augmentation to align the model predictions of the intent detection task. Meanwhile, subword sampling augmentation is used to align the model predictions of both intent detection and slot filling tasks. 
The proposed method consistently improves the SLU performance on the MASSIVE dataset under both full-dataset and zero-shot settings. 
It is worth mentioning that our system ranked 1st in the MMNLU-22 competition under the full-dataset setting. We achieved an exact match accuracy of 49.65 points, outperforming the 2nd system by 1.02 points.

\section{Background}
\subsection{Task Description}

The task of SLU is that given an utterance with a word sequence $\boldsymbol{x} = (x_1, ..., x_n)$ with length $n$. The model is required to solve two sub-tasks. The intent detection task can be seen as an utterance classification task to decide the intent label $o^I$, and the slot filling task is a sequence labeling task that generates a slot label for each word in the utterance to obtain the slot sequence $\boldsymbol{o}^{S} = (o_1^S, ..., o_n^S)$.

\subsection{Dataset Description}
The MASSIVE dataset is composed of realistic, human-created virtual assistant utterance text spanning 51 languages, 60 intents, 55 slot types, and 18 domains~\citep{fitzgerald2022massive}.
There are 11,514 training utterances for each language. For the full-dataset setting, all training data can be used. For the zero-shot setting, only English training data can be used, yet we can translate them into other languages using commercial translators.
There are 2,033, 2,974, and 3,000 utterances for each language in the development, test, and evaluation set, respectively. The average performance in all languages should be reported under the full-dataset setting. Meanwhile, the average performance in all languages except English should be reported under the zero-shot setting.

\subsection{Related Work}
Pre-trained cross-lingual language models~\citep{xlm,xlmr,infoxlm,xlmalign,xlme,mt5}
encode different languages into universal representations and significantly improve cross-lingual transferability. These models usually consist of a multilingual vocabulary~\citep{xlm,xlmr,mt5,vocap} and a Transformer model~\citep{transformer}.

A simple yet effective way to improve cross-lingual fine-tuning is to populate the training data with cross-lingual data augmentation~\citep{xlmr}. 
\citet{xlda} replace a segment of source language input text with its translation in another language as data augmentation. 
\citet{cosda-ml} randomly replace words in the source-language training example with target-language words using the bilingual dictionaries. Then the model is fine-tuned on the generated code-switched data. 
Instead of directly treating cross-lingual data augmentation as extra training data, \citet{xtune} proposed to better use data augmentations based on consistency regularization.

\section{Method}

Given the input utterance $\boldsymbol{x} = (x_1, ..., x_n)$ with length $n$ and the corresponding intent label $o^{I}$ and slot labels $\boldsymbol{o}^{S} = (o_1^S, ..., ... o_n^S)$ from training corpus $\mathcal{D}$, we define the loss for the two sub-tasks of SLU in our fine-tuning process as:
\begin{align}
\mathcal{L}_{I} =~&\sum_{(\boldsymbol{x},o^{I})\in\mathcal{D}}\text{CE}(f_{I}(\boldsymbol{x}), o^{I}) , \nonumber \\
\mathcal{L}_{S} =~&\sum_{(\boldsymbol{x},\boldsymbol{o}^{S})\in\mathcal{D}}\text{CE}(f_{S}(\boldsymbol{x}), \boldsymbol{o}^{S}) , \nonumber 
\end{align}
where $\mathcal{L}_{I}$ and $\mathcal{L}_{S}$ stand for the intent detection task and the slot filling task, $f_{I}(\cdot)$ and $f_{S}(\cdot)$ denote the model which predicts task-specific probability distributions for the input example $\boldsymbol{x}$, $\text{CE}(\cdot, \cdot)$ denotes cross-entropy loss. 

\subsection{Consistency Regularization}
In order to make better use of data augmentations, we introduce the consistency regularization used in~\citet{xtune}, which encourages consistent predictions for an example and its semantically equivalent augmentation. We apply consistency regularization on intent detection and slot filling tasks, which is defined as follows:
\begin{align}
\mathcal{R}_{I} =~&\sum_{\boldsymbol{x}\in\mathcal{D}}\text{KL}_{\text{S}}(f_{I}(\boldsymbol{x}) {\parallel} f_{I}(\mathcal{A}(\boldsymbol{x}, z))) , \nonumber \\
\mathcal{R}_{S} =~&\sum_{\boldsymbol{x}\in\mathcal{D}}\text{KL}_{\text{S}}(f_{S}(\boldsymbol{x}) {\parallel} f_{S}(\mathcal{A}(\boldsymbol{x}, z))) , \nonumber \\
\text{KL}_{\text{S}}(P{\parallel}Q) =~&\text{KL}(\mathrm{stopgrad}(P){\parallel}Q) + \nonumber \\
& \text{KL}(\mathrm{stopgrad}(Q){\parallel}P) \nonumber
\end{align}
where $\text{KL}_{\text{S}}(\cdot{\parallel}\cdot)$ is the symmertrical Kullback-Leibler divergence, $\mathcal{A}(\boldsymbol{x}, z)$ denotes the augmented version of input utterance $\boldsymbol{x}$ with data augmentation strategy $z$.
The regularizer encourages the predicted distributions of the original training example and its augmented version to agree with each other.
The $\mathrm{stopgrad}(\cdot)$ operation\footnote{Implemented by \texttt{.detach()} in PyTorch.} is used to stop back-propagating gradients, which is also employed in~\citep{DBLP:conf/acl/JiangHCLGZ20, DBLP:journals/corr/abs-2004-08994, xtune}.

\begin{figure*}[t]
    \centering
    \includegraphics[scale=0.58]{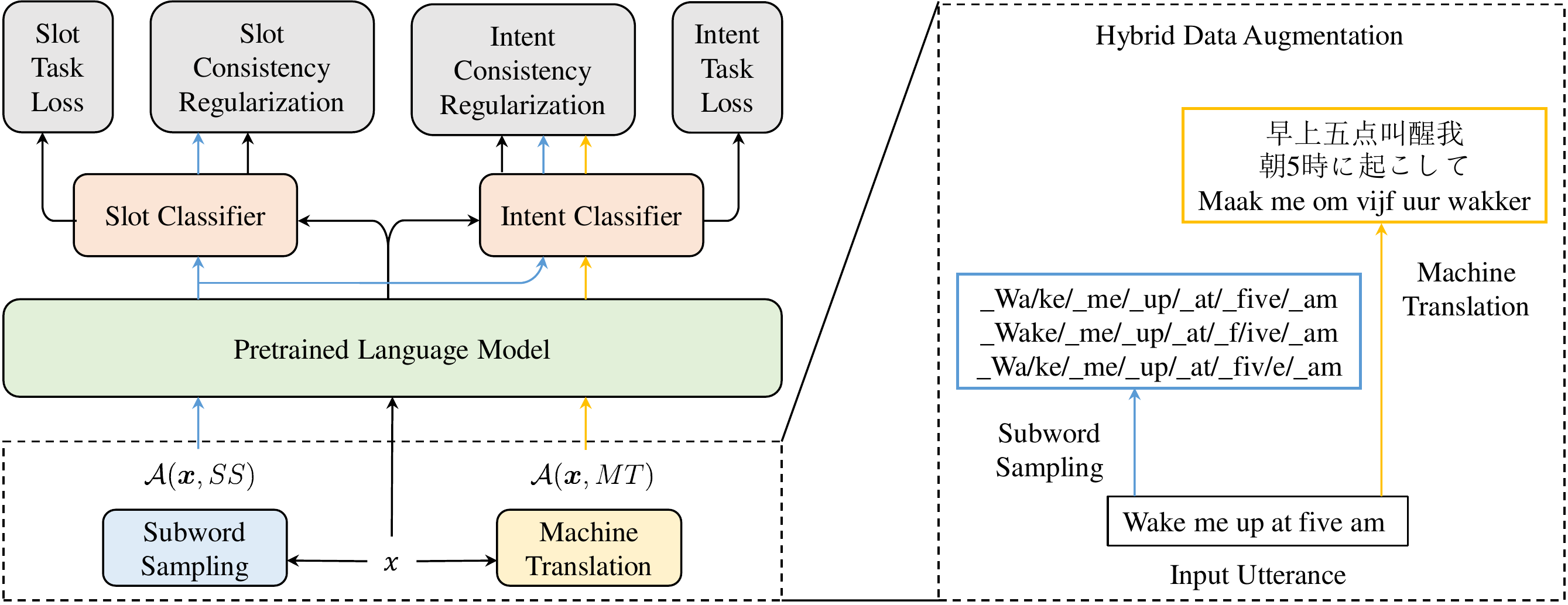}
    \caption{Illustration of our fine-tuning framework. `MT' denotes machine translation augmentation and `SS' denotes subword sampling augmentation.}
    \label{fig:framework}
\end{figure*}
\subsection{Data Augmentations}
We consider two types of data augmentation strategies for our consistency regularization method, including subword sampling and machine translation.
\subsubsection{Subword Sampling}
Subword sampling is to generate multiple subword sequences from the original text as data augmentation. 
We apply the on-the-fly subword sampling algorithm from the unigram language model~\cite{DBLP:conf/acl/Kudo18} in SentencePiece~\cite{DBLP:conf/emnlp/KudoR18}. 
The output distributions of slot labels are generated on the first subword of each word in the input utterance. 
Therefore, the subword sampling augmentation can be used to align the output distribution of both intent detection and slot filling tasks.

\begin{CJK}{UTF8}{gbsn}
\begin{table*}[t]
\centering
\scriptsize
\setlength{\tabcolsep}{1.5mm}
\begin{tabular}{ccccc}
\hline
Text Type & Text Content & Slot Translation & Text Translation & Aligned or Not \\ \hline
Plain Text      & Wake me up at five am Friday this week             & \multirow{2}{*}{\begin{tabular}[c]{@{}c@{}}five am: 凌晨五点\\ Friday this week: 本周周五\end{tabular}} & 本周周五凌晨五点叫我起床              & Yes    \\
Text with Slots in Brackets & Wake me up at {[}five am{]} {[}Friday this week{]} &                                                                                                     & 在{[}凌晨五点{]}{[}本周星期五{]}叫醒我 & No    \\ \hline
Plain Text       & set an alarm for two hours from now                & \multirow{2}{*}{\begin{tabular}[c]{@{}c@{}}two hours from now:\\ 从现在起两小时后\end{tabular}}             & 从现在开始设置两个小时的闹钟            & No    \\
Text with Slots in Brackets & set an alarm for {[}two hours from now{]}          &                                                                                                     & 设置{[}从现在起两小时后{]}的闹钟       & Yes    \\ \hline
\end{tabular}
\caption{Examples of aligning slots into machine translations.}
\label{table:MT}
\end{table*}
\end{CJK}

\begin{table*}[t]
\centering
\small
\begin{tabular}{lcccccc}
\toprule
\multirow{2}{*}{\textbf{Model}} & \multicolumn{3}{c}{\textbf{Test Set}}              & \multicolumn{3}{c}{\textbf{Evaluation Set}}              \\
                       & \textbf{Intent Acc} & \textbf{Slot F1} & \textbf{EMA} & \textbf{Intent Acc} & \textbf{Slot F1} & \textbf{EMA} \\ \midrule
XLM-R Base             & 85.10 & 73.60 & 63.69 & - & - & - \\
XLM-Align Base         & \textbf{86.16} & 76.36 & 66.42 & - & - & - \\ 
mT5 Base Text-to-Text  & 85.33 & \textbf{76.77} & \textbf{66.64} & - & - & - \\ \midrule
XLM-Align Base + Ours  & 87.12 & 77.99 & 68.76 & 85.00 & 68.45 & 48.64 \\ 
mT5 Base Text-to-Text + Ours & \textbf{87.60} & \textbf{78.22} & \textbf{69.60} & \textbf{85.10} & \textbf{69.08} & \textbf{49.65}               \\ \bottomrule
\end{tabular}
\caption{Test and evaluation results on the MASSIVE dataset under the full-dataset setting. Results of XLM-R Base and mT5 Base Text-to-Text are taken from~\citet{fitzgerald2022massive}.}
\label{table:full-dataset}
\end{table*}

\begin{table*}[t]
\centering
\small
\begin{tabular}{lcccccc}
\toprule
\multirow{2}{*}{\textbf{Model}} & \multicolumn{3}{c}{\textbf{Test Set}}              & \multicolumn{3}{c}{\textbf{Evaluation Set}}              \\
                       & \textbf{Intent Acc} & \textbf{Slot F1} & \textbf{EMA} & \textbf{Intent Acc} & \textbf{Slot F1} & \textbf{EMA} \\ \midrule
XLM-R Base             & 70.62 & 50.27 & 38.70 & - & - & - \\
XLM-Align Base         & \textbf{68.49} & \textbf{54.69} & \textbf{40.91} & - & - & - \\ 
mT5 Base Text-to-Text  & 62.92 & 44.77 & 34.72 & - & - & - \\ 
\midrule
XLM-Align Base + Ours  & 85.12 & 71.27 & 62.18 & 83.18 & 62.84 & 43.05 \\
XLM-Align Base + Ours + KD & \textbf{85.76} & \textbf{73.55} & \textbf{64.44} & \textbf{83.89} & \textbf{64.60} & \textbf{44.84} \\
mT5 Base Text-to-Text + Ours  & 84.58 & 69.24 & 60.59 & 82.56 & 60.00 & 40.93 \\ \bottomrule
\end{tabular}
\caption{Test and evaluation results on the MASSIVE dataset under the zero-shot setting. Results of XLM-R Base and mT5 Base Text-to-Text are taken from~\citet{fitzgerald2022massive}.}
\label{table:zero-shot}
\end{table*}

\subsubsection{Machine Translation}
Machine translation is a common and effective data augmentation strategy in the cross-lingual scenario~\citep{xlm, xlda}. 
Due to the difficulty of accessing ground-truth labels in translation examples, machine translation can not be an available data augmentation strategy in the slot filling task.
To improve the quality of our translations, we employ a variety of approaches (See Section~\ref{sec:dp}).
Unlike subword sampling, the output distributions of slot labels between the translation pairs can not be aligned. 
Thus, we only use machine translation to align the output distributions of the intent detection task.

\subsection{Consistency Regularization based on Hybrid Data Augmentations}
We illustrate our fine-tuning framework in Figure~\ref{fig:framework}. We propose to use consistency regularization based on a hybrid data augmentation strategy, which includes data augmentation of machine translation and subword sampling. 
During the training process, we perform task fine-tuning and consistency regularization for an input example simultaneously.
Then the final training loss is defined as follows:
\begin{align}
\mathcal{L} =~&\mathcal{L}_{I} + \lambda_1 \mathcal{L}_{S} + \lambda_2 \mathcal{R}_{I} + \lambda_3 \mathcal{R}_{S} \nonumber
\end{align}
where $\lambda_1$ is the slot loss coefficient, $\lambda_2$ and $\lambda_3$ are the corresponding weights of the consistency regularization for two tasks. We sample different data augmentation for the input example with the pre-defined distribution.

\section{Experiments}

\subsection{Experimental Setup}
We consider two types of pre-trained cross-lingual language models, which are encoder-only models and Text-to-Text models.

We use XLM-Align Base~\citep{xlmalign} for the encoder-only model setting. We use a two-layer feed-forward network with a 3,072 hidden size. 
We use the first representation of sentences ``<s>'' for the intent detection task and the first subword of each word for the slot filling task.

We use mT5 Base~\citep{mt5} for the Text-to-Text model setting. 
We follow~\citet{fitzgerald2022massive} to concatenate ``Annotate: '' and the unlabeled input utterance as the input of the encoder, and generate the text concatenation of the intent label and the slot labels as the decoder output. The labels are separated with white spaces and then tokenized into subwords.

We select the model that performs the best on the development dataset to run prediction on the test and evaluation dataset. We mainly select the batch size in $[32, 64, 128, 256]$, dropout rate in $[0.05, 0.1, 0.15]$, and the hyper-parameters used in our proposed method, including slot loss coefficient $\lambda_1$ in $[1, 2, 4]$, weights of consistency regularization $\lambda_2$ and $\lambda_3$ in $[2, 3, 5, 10]$. We select the learning rate in $[5e^{-5}, 8e^{-5}, 1e^{-4}]$ for Text-to-Text models. As for encoder-only models, we select the learning rate in $[4e^{-6}, 6e^{-6}, 8e^{-6}]$.

\begin{table}[t]
\centering
\small
\setlength{\tabcolsep}{1mm}
\begin{tabular}{lccc}
\toprule
\textbf{Model} & \textbf{Intent Acc} & \textbf{Slot F1} & \textbf{EMA} \\ \midrule
XLM-Align Base + Ours & 87.12 & \textbf{77.99} & \textbf{68.76} \\
- Subword Sampling & \textbf{87.50} & 76.08 & 67.40 \\
- Consistency Regularization & 86.16 & 76.32 & 66.57 \\ \bottomrule
\end{tabular}
\caption{Ablation studies on the MASSIVE test dataset under the full-dataset setting.}
\label{table:ablation-full-dataset}
\end{table}

\subsection{Data Processing}
\label{sec:dp}
For the full-dataset setting, we use examples with the same id in different languages as machine translation augmentation in our fine-tuning framework.  
For the zero-shot setting, we translated the entire English training set into 50 languages using commercial translation APIs, such as DeepL translator and Google translator. 
These translations refer to plain text translations and can be used for intent detection training and consistency regularization.

We used two methods to obtain a translated example that aligned at the slot level. 
One is based on the plain text translation. Each slot value in an English training example is translated into a target language. 
If the translation results of each slot can be found in the plain text translation, a slot-aligned translation is obtained.
The other is based on the annotated English training examples. We translated the annotated English training example with brackets for slot values (without slot type in brackets). Using brackets explicitly allows the translator to align slots to consecutive spans. And we also translated each slot value into the target language. If the translation result of each slot can be found in the annotated utterance translation, we obtain a slot alignment example after removing the brackets.

In practice, slot-aligned examples based on plain text translations are preferred as the final result of the slot alignment. If no such example is available, we use the slot-aligned results from annotated translations. Examples of slot alignment are shown in Table~\ref{table:MT}. For those plain text translations where we can not align the slot labels, we only use them for the training of the intent detection task.

\subsection{Evaluation Metrics}
The evaluation in competition is mainly conducted using three metrics:
\begin{itemize}
    \item Exact Match Accuracy (EMA): The percentage of utterance-level predictions where the intent and all slots are exactly correct.
    \item Intent Accuracy (Intent Acc): The percentage of predictions in which the intent is correct.
    \item Slot Micro F1 (Slot F1): The micro-averaged F1 score is calculated over all slots.
\end{itemize}

\subsection{Results}
Table~\ref{table:full-dataset} shows our results on the MASSIVE dataset under the full-set setting. 
We tried different cross-lingual pre-trained language models under the baseline setting.
Among them, XLM-Align Base performs the best on the intent detection task, while the mT5 Base Text-to-Text model performs the best on the slot filling task and exact match accuracy. 
When applying our consistency regularization method, the mT5 Base Text-to-Text model outperforms the XLM-Align Base model by 0.84 points and 0.99 points on exact match accuracy on the test dataset and the evaluation set, respectively. 
Meanwhile, compared to the baseline model, using consistency regularization achieves an absolute 2.96-point improvement on exact match accuracy with the mT5 Base Text-to-Text model.

Table~\ref{table:zero-shot} shows our results on the MASSIVE dataset under the zero-shot setting. 
For the baseline models, XLM-Align Base performs the best on all three metrics. 
Difference from the full-dataset setting, mT5 Base Text-to-Text models perform poorly under the zero-shot setting. 
We attribute it to the fact that Text-to-Text models strongly rely on the training data quality since most of the training data under the zero-shot setting are obtained with machine translation systems. 
When applying our consistency regularization method, the XLM-Align Base model outperforms the baseline model by 21.27 points. 
Distilled from the InfoXLM Large~\citep{infoxlm} model will further improve the performance by an absolute 2.26-point.

\subsection{Ablation Studies}
We conduct ablation studies on the test dataset of MASSIVE under the two settings. Table~\ref{table:ablation-full-dataset} shows the results under the full-dataset setting. Ablating subword sampling will degrade the performance by 1.36 points on the exact match accuracy, where the performance drop comes mainly from the slot filling task, indicating the subword sampling augmentation mainly works on slot filling. 
Ablating consistency regularization will degrade the performance by 2.19 points on the exact match accuracy. 
The performances on both intent detection and slot filling tasks are decreased.

The zero-shot setting results are presented in Table~\ref{table:ablation-zero-shot}. It can be observed that when machine translation augmentation is removed, the exact match accuracy drops by 16.68 points, while the performance on intent detection and slot filling are also significantly worse. 
We also removed the subword sampling augmentation, and the performance is found to have the same trend as in the full-dataset setting. An absolute 1.24-point drop on the exact match accuracy and an absolute 1.75-point drop on slot micro F1 demonstrate that subword sampling is more beneficial for the slot filling task.
By removing the consistency regularization, the performance of exact match accuracy will degrade by 2.23 points. The performance shows a significant performance drop on both intent detection and slot filling tasks.

\begin{table}[t]
\centering
\small
\setlength{\tabcolsep}{1mm}
\begin{tabular}{lccc}
\toprule
\textbf{Model} & \textbf{Intent Acc} & \textbf{Slot F1} & \textbf{EMA} \\ \midrule
XLM-Align Base + Ours        & 85.12 & \textbf{71.27} & \textbf{62.18} \\
- Subword Sampling           & \textbf{85.14} & 69.52 & 60.94 \\
- Machine Translation        & 72.27 & 58.37 & 45.50 \\
- Consistency Regularization & 83.90 & 69.37 & 59.95 \\ \bottomrule
\end{tabular}
\caption{Ablation studies on the MASSIVE test dataset under the zero-shot setting.}
\label{table:ablation-zero-shot}
\end{table}

\section{Conclusion}
We propose to use consistency regularization based on a hybrid data augmentation strategy to improve the performance of multilingual SLU. 
The proposed method is flexible and can be easily plugged into the fine-tuning process of both the encoder-only model and the Text-to-Text model. 
The experimental results demonstrate the importance of consistency regularization and the hybrid data augmentation strategy, respectively.

\section*{Acknowledgments}

This work was supported by the National Key R\&D Program of China via grant 2020AAA0106501 and the National Natural Science Foundation of China (NSFC) via grant 62236004 and 61976072.

\bibliography{anthology,custom}
\bibliographystyle{acl_natbib}

\appendix

\end{document}